\documentclass{article}

\usepackage[nonatbib, final]{neurips_2021}

\usepackage[utf8]{inputenc} 
\usepackage[T1]{fontenc}    
\usepackage{url}            
\usepackage{booktabs}       
\usepackage{amsmath}
\usepackage{amssymb}
\usepackage{amsfonts} 
\usepackage{amsthm}
\usepackage{algorithm}
\usepackage{caption}
\usepackage{subcaption}
\usepackage{algpseudocode}
\usepackage{authblk}
\usepackage{nicefrac}       
\usepackage{microtype}      
\usepackage[dvipsnames]{xcolor}         
\usepackage{graphicx}
\usepackage{multicol}
\usepackage{svg}
\graphicspath{{figures/}}

\usepackage[numbers]{natbib}

\usepackage{hyperref}
\hypersetup{colorlinks,allcolors=NavyBlue}

\newcommand{\E}{\mathbb{E}}

\newcommand{\Prob}{\mathbb{P}}
\DeclareMathOperator{\testEdge}{stitch}
\DeclareMathOperator{\getPotentialEdges}{getCandidateEdges}
\DeclareMathOperator{\getNeighbors}{getNeighbors}
\DeclareMathOperator{\getFutureEdges}{getFutureEdges}
\DeclareMathOperator{\CEM}{CEM}

\DeclareMathOperator{\argmax}{arg max}
\newtheorem{theorem}{Theorem}

\newtheorem{corollary}{Corollary}
\newtheorem{definition}[]{Definition}
\newcommand{\Statespace}{\mathcal{S}}
\newcommand{\onpolbisim}{\sim_\pi}
\newcommand{\Actionspace}{\mathcal{A}}
\newcommand{\norm}[1]{\left\lVert#1\right\rVert}

\newcommand*\samethanks[1][\value{footnote}]{\footnotemark[#1]}

\title{BATS: Best Action Trajectory Stitching}

\author[ ]{\textbf{Ian Char$^1$}\thanks{Equal contribution author.}}
\author[ ]{\textbf{Viraj Mehta$^2$}\samethanks[1]}
\author[2]{\textbf{Adam Villaflor}}
\author[2]{\textbf{John M. Dolan}}
\author[1]{\textbf{Jeff Schneider}}
\affil[1]{Department of Machine Learning, Carnegie Mellon University}
\affil[2]{Robotics Institute, Carnegie Mellon University}
\affil[ ]{\textit{\{ichar, virajm, avillaflor, jdolan, schneide\}@cs.cmu.edu}}
\date{}                     
\begin{document}

\maketitle

\begin{abstract}


The problem of offline reinforcement learning focuses on learning a good policy from a log of environment interactions. Past efforts for developing algorithms in this area have revolved around introducing constraints to online reinforcement learning algorithms to ensure the actions of the learned policy are constrained to the logged data. In this work, we explore an alternative approach by planning on the fixed dataset directly. Specifically, we introduce an algorithm which forms a tabular Markov Decision Process (MDP) over the logged data by adding new transitions to the dataset. We do this by using learned dynamics models to plan short trajectories between states. Since exact value iteration can be performed on this constructed MDP, it becomes easy to identify which trajectories are advantageous to add to the MDP. Crucially, since most transitions in this MDP come from the logged data, trajectories from the MDP can be rolled out for long periods with confidence. We prove that this property allows one to make upper and lower bounds on the value function up to appropriate distance metrics. Finally, we demonstrate empirically how algorithms that uniformly constrain the learned policy to the entire dataset can result in unwanted behavior, and we show an example in which simply behavior cloning the optimal policy of the MDP created by our algorithm avoids this problem.
\end{abstract}


\section{Introduction}
The goal of Reinforcement Learning (RL) is to learn a policy which makes optimal actions for a decision making problem or control task. The field of \textit{deep} RL, in which one learns neural network models to represent key quantities for decision making, has recently made great strides \citep{alphago, mnih2013atari, rl_robotics}. In many deep RL algorithms, this involves learning a neural network for both the policy and the value function, which estimates the value of states or state-action-pairs with respect to the current policy. Many promising model-based methods \citep{PETS, mbpo} also learn a deep dynamics function that estimates next states given current states and actions. 

In the standard, \textit{online} setting, the policy is repeatedly deployed in the environment during training time, which provides a continual stream of on-policy data that stabilizes the learning procedure. However, the \textit{online} setting is unreasonable for applications, since it requires a way to cheaply and safely gather a large number of on-policy samples. As such, there has been increasing interest in the so-called \textit{offline} setting \citep{levine2020offline} in which a policy is learned solely from logged off-policy data. 

However, the offline setting comes with its own problems. Simply applying deep reinforcement learning algorithms designed for the online setting will often cause exploding value estimates because of distribution mismatch and recursive updates \citep{kumar2019stabilizing}. In model-based methods, the combination of small initial errors and test-time distribution shift often leads to rapidly accumulating model error. While distribution shift and model exploitation are potential issues in online RL, these problems are more severe in the offline setting, as the agent cannot collect additional experience to rectify compounding errors in estimation or planning.  To address these problems, offline RL algorithms add constraints to encourage the agent to only operate in the support of the data by either constraining the policy \cite{wu2019behavior, kumar2019stabilizing} or penalizing uncertain state-actions \cite{yu2020mopo, morel, kumar2020conservative, yu2021combo}.

 Rather than trying to implicitly constrain the agent to stay in the support of the data,
 in this work we explore what happens if we plan over the logged data directly. In particular, we create a tabular MDP by planning short trajectories between states in the dataset, and then we do exact value iteration on this MDP. Unlike other model-based methods which are limited to short imagined trajectories, trajectories from our MDP are mostly comprised of real transitions from the dataset and can therefore be rolled out for much longer with confidence. As such, we argue that our algorithm is able to better reason about the dataset as a whole. In this work, we show that re-imagining the offline dataset in this way allows for the following:
\begin{itemize}
    \item By coupling together long trajectories with exact value iteration, our algorithm is able to better estimate the resulting policy's value. We prove that under the correct distance metrics our algorithm can be used to form upper and lower bounds for the value function. We demonstrate empirically that this aligns well with the value of a policy behavior cloned on these trajectories.
    \item By performing full rollouts in our tabular MDP, we are able to approximate our optimal policy's occupancy distribution. We show how many algorithms that uniformly constrain the learned policy to actions on the dataset struggle with ``undirected'' datasets (i.e., data collected without a specific reward function in mind) and demonstrate that our algorithm avoids this problem by filtering out data unrelated to the task.
\end{itemize}

\section{Preliminaries}
\label{ss:preliminaries}

In this work, we assume the environment can be represented as a \textit{deterministic}, infinite horizon MDP $M = \langle \Statespace, \Actionspace, \gamma, T, r, \rho\rangle$, where $\Statespace$ is the state space, $\Actionspace$ is the action space, $\gamma \in (0, 1)$ is the discount factor, $T: \Statespace \times \Actionspace \to \Statespace$ is the transition function, $r: \Statespace \times \Actionspace \to \mathbb{R}$ is the reward function, and $\rho$ is the initial state distribution. We refer to an MDP as \textit{tabular} if it has a finite state and action space. While we assume that the true environment in question has every action available to be played at every state (i.e. $T(s, a)$ is well-defined $\forall s \in \Statespace, a \in \Actionspace$), later in this work we also consider MDPs that have only a subset of actions for each state. When this is the case, we denote $\Actionspace^s \subset \Actionspace$ to be the actions available to be played at state $s \in \Statespace$. Such MDPs are defined as $M = \langle \Statespace, \{\Actionspace^s\}_{s \in \Statespace}, \gamma, T, r, \rho\rangle$.

In reinforcement learning, we attempt to learn a stochastic policy $\pi(a  | s): \Statespace \to P(\Actionspace)$, where $P(\Actionspace)$ is the set of all distributions over $\Actionspace$. We desire our learned policy to maximize the expected discounted sum of rewards,
$\E_{\pi, \rho} \left [ \sum_{t=0}^{\infty} \gamma^t r(s_t, a_t) \right ]$
, where $s_t = T(s_{t - 1}, a_{t - 1})$, $a_t \sim \pi(\cdot | s_t)$, and $s_0 \sim \rho$. To facilitate the optimization of this quantity, we can define an optimal state-action value function $Q^{*}: \Statespace \times \Actionspace \to \mathbb{R}$ that satisfies the following recurrence relation known as the Bellman Equation: 
\begin{align}
    Q^{*}(s, a) &= r(s, a) + \gamma \max_{a'} Q^{*}(T(s, a), a')
\end{align}
Then, we can reformulate policy optimization as trying to solve $\pi(s) = \underset{a}{\argmax}  Q^{*}(s, a)$,  $\forall s \in \Statespace$, where we can estimate $Q^{*}$ by iteratively performing the bellman update to $Q_{k+1}(s, a) \gets r(s, a) + \gamma \underset{a'}{\max} Q_{k}(T(s, a), a')$. In tabular MDPs, the Q-function and its updates can be written and performed exactly. Thus, this procedure -- known as \textit{value iteration} -- will eventually converge, i.e. $Q_k(s, a) \to Q^{*}(s, a)$,  $\forall (s, a) \in \Statespace \times \Actionspace$ as $k \to \infty$. However, in general MDPs where there is a possibly infinite number of states or actions, we must rely on function approximation and finite samples to instead perform \textit{approximate value iteration}, which is not guaranteed to converge. For notational convenience, we denote the value of a state as $V^*(s) = \max_a Q^*(s, a)$. Policy $\pi$'s \emph{occupancy distribution} is defined as $\mu_\pi(s) \propto \sum_{i=0}^\infty \gamma^{i}p(s_i=s)$, where $s_i = T(s_{i - 1}, a_{i - 1})$, $a_{i-1} \sim \pi(\cdot | s_{i - 1})$, and $p(s_0) \equiv \rho$. We denote the value function for $\pi$ as $V^{\pi}$; that is, $V^{\pi}(s)$ is the expected, cumulative discounted sum of rewards from playing policy $\pi$ starting from state $s$. When it is not clear from context, we denote $V^\pi_M$ as the value for the function $\pi$, specifically over MDP $M$.

In offline reinforcement learning, one assumes access to a fixed set of environment interactions. In this work, we assume that we have access to a dataset, $D = \bigcup_{j \in [N]}\{(s_{ji}, a_{ji}, s'_{ji}, r_{ji})\}_{i=1}^{t_j}$, which is comprised of $N$ trajectories of possibly varying lengths, $t_j$. For the remainder of Section~\ref{ss:preliminaries}, we use $s_{ji}, s'_{ji}, a_{ji}, r_{ji}$ to represent the current state, next state, action played, and reward received for the $i^{th}$ timestep of the $j^{th}$ trajectory. Also note that, if $i < t_j$, then $s_{j(i+1)} = s'_{ji}$.


\textbf{Structures over the Offline Data.} Given a dataset, $D$, collected in MDP $M = \langle \Statespace, \Actionspace, \gamma, T, r, \rho\rangle$, one can construct a tabular MDP that incorporates only the states and actions observed in the dataset. We denote this MDP as $M_0 = \langle \Statespace_0, \{\Actionspace^s_0\}_{s \in \Statespace_0}, \gamma, T_0, r_0, \rho_0\rangle$, where $\Statespace = \cup_{j \in [N], i \in [t_j]} \left( \{s_{ji}\} \cup \{s'_{ji}\} \right)$, $\Actionspace_0^s = \{a_{ji} | \forall j \in [N], \forall i \in [t_j] \textrm{ s.t. } s_{ji} = s \}$, $T_0(s, a) = T(s, a)$, $r_0(s, a) = r(s, a)$, and $\rho_0$ is a discrete uniform distribution over $\{s_{j0}\}_{j=1}^M$. 

It will often be beneficial to describe the offline dataset from a graphical perspective. A graph, $G := (V, E)$, is fully characterized by its vertex set, $V$, and its edge set, $E$. We note that the notation for the vertex set is overloaded with the value function, but the difference is often clear from context. For any MDP, we can define a corresponding graph that has a vertex set which is the same as the MDP's state space and an edge set which matches the MDP's transition function. For example, the graph corresponding to $M_0$,  $G_0 = (V_0, E_0)$, has vertex set, $V_0 = \Statespace_0$, and edge set, $E_0 = \{(s, T_0(s, a)) | s \in \Statespace_0, a \in \Actionspace_0^s\}$. Specific to this paper, we also consider the undirected, \textit{neighbor graph}, $G_\epsilon$, which has the same vertex set, but has edge set such that $\{s, s'\}$ is an edge iff $\norm{s - s'} \leq \epsilon$ for a specified norm and $\epsilon > 0$.

\textbf{Bisimulation Metric.} In this work, we use the \emph{on-policy bisimulation distance} from \citet{castro2019scalable}, which also gives a sampling-based algorithm for approximating such a metric. We denote this as $d_\sim^\pi(\cdot, \cdot)$. A key result about this metric is the following:
\begin{theorem}[Theorem 3 from \citet{castro2019scalable}]
\label{thm:bisimulation}
Given states $s, t \in \Statespace$ in an MDP, $M$, and a policy, $\pi$,
\begin{align*}
|V^\pi(s) - V^\pi(t)| \leq d_\sim^\pi(s, t).    
\end{align*}
\end{theorem}
In other words, $d_\sim^\pi(\cdot, \cdot)$ is a metric over states for which the value function is $1$-Lipschitz continuous. We discuss bisimulation further in Appendix \ref{a:bisimulation}.

\section{Method}

The MDP, $M_0$, as described in Section~\ref{ss:preliminaries}, has several desirable properties. First, it is tabular, so one can easily apply value iteration to find the optimal policy. Second, policies defined over $M_0$ will be conservative since policies can only choose actions that appear in the dataset. Unfortunately, $M_0$ is so conservative that it is uninteresting since there is little to no choice in what actions can be made at each state. We must make additions in order to create a more interesting MDP to optimize. Our solution is to create transitions via planning: an operation we call \textit{stitching}.

\subsection{The Stitching Operation}

Broadly speaking, the so-called stitching operation simply adds a transition (or sequence of transitions) from one pre-existing state in an MDP to another pre-existing state via planning in a learned model. To flesh this operation out, suppose that, for a tabular MDP, $\hat{M} = \langle \hat{S}, \{\hat{\Actionspace}^s\}_{s \in \hat{\Statespace}}, \gamma, \hat{T}, \hat{r}, \rho_0 \rangle$, we would like to add the ability to transition from $s \in \hat{\Statespace}$ to $s' \in \hat{\Statespace}$. Here, $\hat{M}$ is either $M_0$, or $M_0$ with some additional states, actions, and transitions included. Using a learned dynamics model, $\tilde{T}$, as a proxy to the the true dynamics, $T$, we can find actions that transition from $s$ to $s'$ via \textit{planning}, i.e. we can solve the following optimization problem:
\begin{align} \label{eq:planning} 
    \underset{a_0, \ldots, a_{k -1} \in \Actionspace}{\textrm{argmin}} \norm{s' - s_k} \quad \textrm{ where } s_j = \tilde{T}(s_{j-1}, a_{j - 1}),  \forall j=1,\ldots, k \quad \textrm{ and } s_0 = s
\end{align}
where $k$ is the number of actions allowed and $\Actionspace$ is the set of actions available in the environment. We choose to optimize this objective with the Cross Entropy Method (CEM), as used in \citet{PETS}. For a specified tolerance, $\delta \in \mathbb{R}$, we consider it possible to transition from $s$ to $s'$ if there exists a solution such that $\norm{s' - s_k} < \delta$. If the tolerance cannot be achieved, we leave $\hat{M}$ unchanged. Otherwise, we set $\hat{\Statespace} = \hat{\Statespace} \cup \{ s_i\}, \hat{\Actionspace}^{s_i} = \hat{\Actionspace}^{s_i} \cup \{a_i\}$ for $i = 0,\ldots, k - 1$, where $\hat{\Actionspace}^{s_i} = \emptyset$ if $s_{i - 1} \notin \hat{\Statespace}$. If $i < k - 1$, we set , $\hat{T}(s_i, a_i) = s_{i + 1}$, and otherwise set $\hat{T}(s_{k - 1}, a_{k -1}) = s'$. Lastly, $\hat{r}(s_i, a_i) = \tilde{r}(s_i, a_i) - c d(s_k, s')$ for $i = 0, \ldots, k - 1$, where $\tilde{r}$ is a learned estimate of the reward function, $c$ is a penalty coefficient, and $d$ is an appropriate distance metric. The addition of the penalty term encourages policies to choose transitions that occur in the dataset over the possibly erroneous ones that are added via stitching. Choosing $d$ to be a bisimulation distance has theoretical ramifications which we will discuss in Section~\ref{ss:extendo}.


\subsection{The BATS Algorithm}

\begin{figure}[t!]
     \centering
     \includegraphics[width=0.7\textwidth]{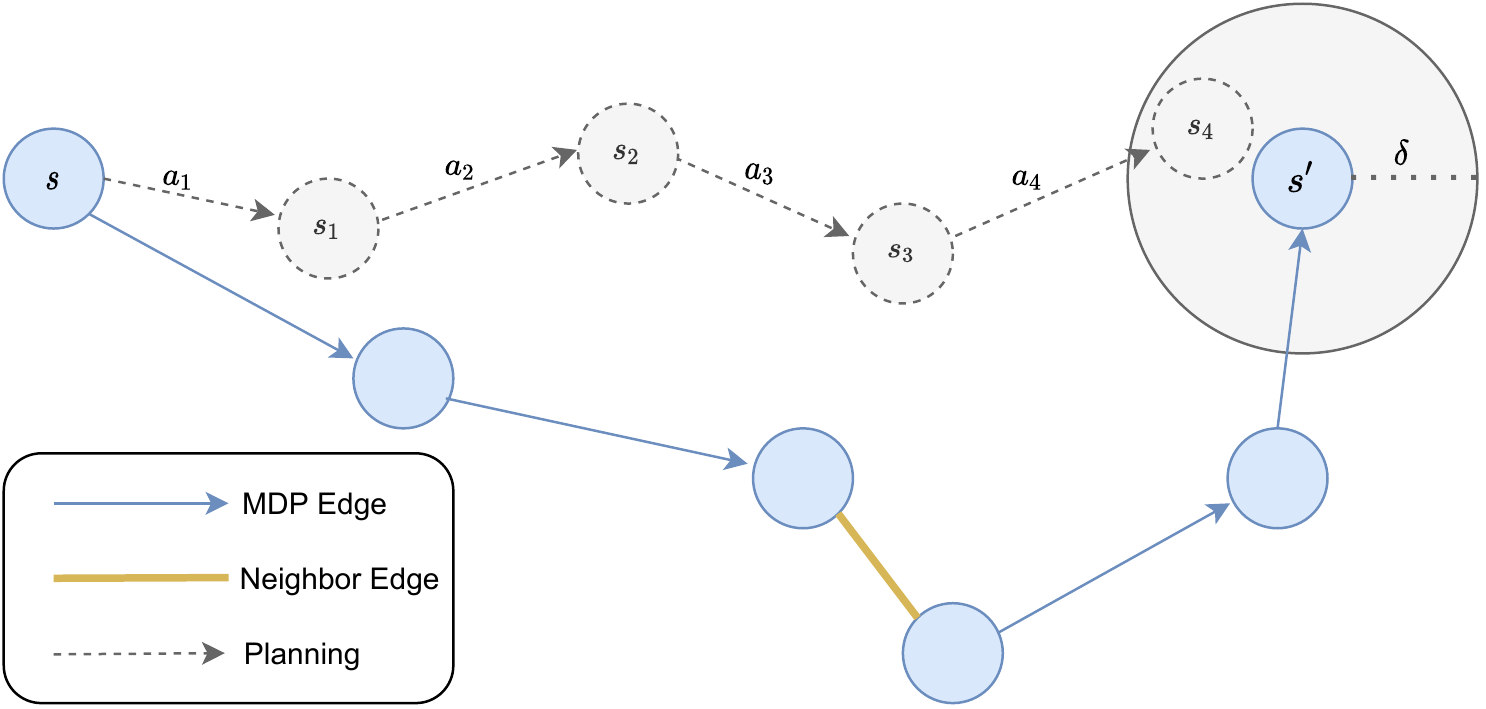}
     \label{fig:stitch_cartoon}
    \caption{\textbf{A diagram of the stitching procedure in BATS.} The blue edges come from the directed graph, $G_i$, the yellow edge comes from the neighbor graph, $G_{\epsilon}$, and dashed gray edges are the planned actions. Here, the $(s, s')$ state pair is a viable candidate to try to stitch with $k=4$ actions because there are 4 blue edges and one yellow edge forming a path from $s$ to $s'$. This would be considered a successful stitch since $s_4$ is within $\delta$ of $s'$. \vspace{-5mm}}
    \label{fig:cartoon}
\end{figure}

Given unlimited compute, the ideal algorithm would be to attempt to stitch all pairs of states in the graph and then perform value iteration on the resulting tabular MDP.
However this is not often feasible, which is where our algorithm, \textit{Best Action Trajectory Stitching (BATS)}, comes into play. BATS is an iterative algorithm where, for each $i = 0, 1, \ldots, n - 1$, we perform value iteration for MDP, $M_i$, to find optimal policy $\pi_i$, we narrow the pool of candidate stitches to those that are both feasible and impactful, 
and lastly we run the stitching procedure over $M_i$ and set the results as $M_{i + 1}$.
We will now discuss the heuristics for identifying feasible and impactful stitches. The full algorithm is written out concretely in Appendix~\ref{s:subroutines}.

\textbf{Identifying Feasible Stitches.} We first establish a notion of \textit{feasible} stitches.
We operate on the following heuristic: if there exists a sequence of actions that lead to $s'$ from $s$, there is likely a similar sequence of actions that lead to $s'$ starting at a state neighboring $s$. Concretely, for iteration $i$, we only consider stitching a state, $s$, to another state, $s'$, if there exists a path from $s$ to $s'$ that uses at most $K$ edges from graph $G_i$ (i.e. the graph corresponding to $M_i$) and exactly one edge from the nearest neighbor graph $G_\epsilon$ (this is visualized in Figure~\ref{fig:cartoon}). If we find that $k$ edges from $G_i$ are used in the path from $s$ to $s'$, we limit the planning procedure in the stitching operation to optimize for $k$ actions. To introduce more conservatism, we also only consider $s' \in \Statespace_0$; that is, we do not consider stitching to any ``imagined'' states that may be the result of previous stitching. This constraint enforces the agent to stay in distribution. 

\textbf{Identifying Impactful Stitches.} To help identify impactful stitches during iteration $i$, we focus on making stitches that maximizes $\E_{s \sim \mu_{\pi_{i + 1}}} \left[ V^{\pi_{i + 1}}_{M_{i + 1}} (s) \right]$. To do this heuristically, we first sample $s_1, \ldots, s_m$ from $\mu_{\pi_i}$ and find all feasible destinations each sample could stitch to. Let $s$ be one such sample, and let $s'$ be a feasible destination state. Suppose there is a path connecting these states that uses exactly $k$ edges from $G_i$ and one edge from $G_\epsilon$. Let $s'_k$ be the state that $\pi_i$ transitions to after acting $k$ times in $M_i$. Then, we consider $(s, s')$ to be a candidate stitch if $V^{\pi_i}(s') > V^{\pi_i}(s'_k)$. In other words, if it is believed that $s'$ can be reached from $s$ in $k$ transitions, then the stitch between $s$ and $s'$ only deserves our attention if $s'$ is more valuable than the state that $\pi_i$ currently transitions to in $k$ steps.

After running the BATS algorithm for $n$ iterations, we are left with an optimal policy, $\pi_n$ for the stitched, tabular MDP, $M_n$; however, this policy cannot be deployed on the true, continuous MDP, $M$, since the domain of $\pi_n$ is only a subset of $\Statespace$. To remedy this, we collect a large number of trajectories using $\pi_n$ in $M_n$ to generate a dataset of state-action tuples to train a parametric policy with behavioral cloning. However, we note that alternative policy learning algorithms could be used to make a policy well-defined over $\Statespace$.

\textbf{Hyperparameters.} Our algorithm has a number of hyperparameters of interest. Dynamics model training, metric learning, behavior cloning, and CEM all have parameters which trade off computation and performance. However, these are well-known methods which operate independently and for which hyperparameter selection is a relatively well-understood problem. The BATS algorithm itself requires a tolerance for value iteration, $\epsilon$ for the neighbors graph, $\delta$ for planning tolerance, $m$ for the number of samples from occupancy distribution per iteration, $K$ for the max number of actions in a stitch, and $n$ for the number of iterations. We further discuss how we determined hyperparameters for our experiments in the Appendix~\ref{app:hyperparameters}.

\vspace{-2mm}

\section{Analyzing BATS with Bisimulation Distance}\label{ss:extendo}

\paragraph{Assumptions}
Let $M_0$ be a tabular MDP formed from an offline dataset collected in $M$, as previously formulated. We can extend $\pi$ to the domain $\Statespace$ by behavior cloning; that is, by finding the member of a parameterized policy class which has minimum training error regularized with weight norm. For a slight simplification of our analysis, we assume that the hypothesis class is rich enough that it can interpolate the training set; that is, it can reach zero behavior cloning error for all $s \in \hat{\Statespace}$. We often refer to both a policy and its extension as $\pi$.

We also assume that on a finite MDP an optimal policy can be found efficiently, and that the learned dynamics model, $\tilde{T}$, is accurate for short transitions in the support of the data. Although in practice, we will learn the reward function, in this analysis we also assume the reward function, $r$, is known. Lastly, we assume that we are able to learn an embedding, $\phi^\pi: \Statespace\to Z$, such that the $L2$ norm in the latent space, $Z$, is the on-policy bisimulation metric. That is, we can learn a $\phi^\pi$ such that if $||\phi^\pi(s) - \phi^\pi(s')|| < \epsilon$ then $|V^\pi(s) - V^\pi(s')| < \epsilon$.

\paragraph{Sandwich Bound on Value}

Consider the collection of $\ell$ tuples $\{(b_j, c_j, a_j)\}_{j = 1}^\ell$, where $b_j, c_j \in \Statespace_0$, $a_j \in \Actionspace$, and $a_j \notin \Actionspace_0^{b_j}$ for all $j \in [\ell]$.
Then define $M^-$ as the MDP derived by starting from $M_0$ and, for each $j \in [\ell]$, setting $\Actionspace_0^{b_j} = \{a_j\} \cup \Actionspace_0^{b_j}$, $T_0(b_j, a_j) = c_j$, and $r_0(b_j, a_j) = r(b_j, a_j) - \gamma \epsilon_j$, where $\epsilon_j > 0$ is some notion of penalty. In other words, $M^-$ is the result of making $\ell$ stitches in $M_0$ where $K = 1$. There exists a policy $\pi^-$ which is the optimal policy on $M^-$ and extends by behavior cloning to $\Statespace$. Similarly, we can construct MDP $M^+$ in the exact same way as $M^-$, but by setting reward to $\hat{r}(b_j, a_j) = r(b_j, a_j) + \gamma \epsilon_j$ for each $j \in [\ell]$. In this setting we can bound the value of $\pi^-$ in the true MDP on either side by its value attained in the finite MDPs just defined. We formalize this notion in the following Theorem.

\begin{theorem}
\label{thm:value_function_bound}
For $j \in [\ell]$, let each penalty term, $\epsilon_j$, be such that $||\phi^{\pi^-}(T(b_i, a_i)) - \phi^{\pi^-}(c_i)|| < \epsilon_j$.

Then
\[\forall s \in \Statespace_0, V_{M^-}^{\pi^-}(s) \leq V^{\pi^-}_{M}(s) \leq V_{M^+}^{\pi^-}(s) .\]
\end{theorem}

In other words, under the correct assumptions, we can construct a pessimistic and optimistic MDP. The value of policy $\pi^-$ in the pessimistic and optimistic MDP will bound the true value of this policy from below and above, respectively. We give the proof in Appendix~\ref{ss:value_function_proof} and a short sketch here: The value function can be written as $V_M^{\pi^-}(s) = \sum_{s_i}\gamma^i r(s_i, \pi(s_i))$, which can be lower and upper-bounded using Theorem \ref{thm:bisimulation} for every transition in the expansion which does not exist in the dataset. This is accomplished by taking into account a pessimistic planning error of the dynamics model. Although Theorem~\ref{thm:value_function_bound} is for the case where we limit stitches to have at most one action, it is likely easy to extend this result to more actions.

There are 3 major implications of the theorem. First, if the behavior cloning, dynamics model, and bisimulation assumptions hold, the value function estimates must be accurate. Second, reasoning by contraposition gives that if the value function estimates are bad, it is due to errors in these components. As such, one should recompute the edge penalties as the policy changes by fine-tuning the bisimulation metric. Third, if the current lower bound is higher than a previous upper bound on the value, the policy is guaranteed to have improved.

We formalize the third fact in the following corollary. Starting with the setup from before, let $M'^-$ and $M'^+$ be tabular MDPs constructed using the alternative sequence of tuples $\{(b_j', c_j', a_j')\}_{j = 1}^{\ell'}$. Let $\epsilon_j'$ be the penalty term used in formulating these MDPs, and let $\pi'^-$ be the optimal policy for $M'^-$.

\begin{corollary}
\label{cor:policy_improvement}

Let $\epsilon_j$ and $\epsilon_j'$ satisfy the assumptions of Theorem~\ref{thm:value_function_bound} for mappings $\phi^{\pi^-}$ and $\phi^{\pi'^-}$, respectively. 
If for some $s \in \Statespace_0$, $V_{M^+}^{\pi^-}(s) <V_{M'^-}^{\pi'^-}(s)$, then $ V^{\pi^-}_{M}(s) < V_{M}^{\pi'^-}(s)$.
\end{corollary}
This corollary is a natural consequence of Theorem~\ref{thm:value_function_bound} and implies the policy $\hat{\pi}^-$ is better at state $s$ than $\pi^-$. If this holds on average for $s \sim \rho$, then we can conclude that $\hat{\pi}^-$ is the better policy.

\vspace{-3mm}
\section{Illustrative Example: Mountain Car}
\vspace{-3mm}
\label{sec:mc}

\begin{figure}
    \centering
    \includegraphics[width=\textwidth]{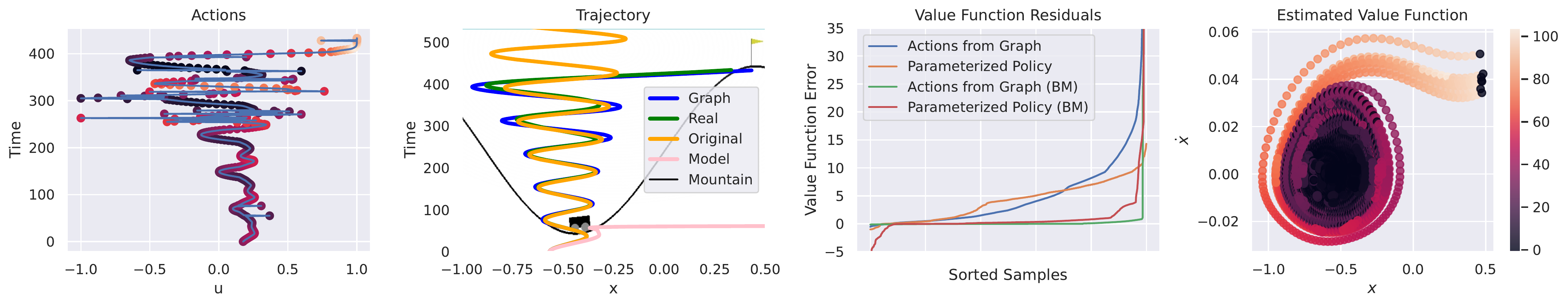}
    \caption{\textbf{Mountain Car Example}. The plots show example actions, associated trajectories, the value function lower bound, and residuals for the standard dynamics model and the bisimulation metric. \vspace{-6mm}}
    \label{fig:mc}
\end{figure}

As an initial test of the algorithm and analysis, we trained BATS on a small dataset consisting mostly of random actions along with 5 expert trajectories on a continuous mountain car environment.
Behavior cloning fails on this dataset, but running BATS and subsequently behavior cloning trajectories from the stitched MDP solves the problem, reliably getting returns of 95 (90 is considered solved). In Figure \ref{fig:mc}, we show how this happens. Starting at the left, the actions are in general concomitant with a good mountain car policy but with occasional spikes.
The spikes result from the planning step in BATS, which has the objective solely to reach the next intended state as accurately as possible. Although the large control inputs are costly in this problem, they are only intermittently added and in general result in a policy that solves the problem. 

The second panel makes clear how BATS leverages real transitions to produce realistic, trustworthy trajectories. To show this, we show a stitched trajectory (blue) that was originally (gold) unable to reach the goal. Replaying the actions from this trajectory in the real environment, we find that the trajectory closely matches what happens in the environment (green). However, replaying the same actions in our learned dynamics model results in horrendous error before 100 timesteps (pink). This demonstrates how our method can produce more effective novel long-term trajectories than purely relying on rolling out a learned model. 


The third panel shows the error in the graph value function estimates over a sampling of states from the graph from the true returns experienced both by executing the actions associated with the edges taken in the graph and by executing a policy cloned from the graph. We also train a bisimulation model following work from \citet{zhang_bisim, castro2019scalable} and execute BATS according to this metric (see Appendix~\ref{app:bisim_implementation}). We find that doing this results in the value function estimates that are quite accurate. One interesting feature is that on the left side, there are actually states where the cloned policy does \emph{better} than the graph policy. We believe this is likely due to the cloned policy smoothing out some of the control spikes output by the planning process and using less control input. This panel admits a natural decomposition of the errors in BATS. The small errors in executing the bisimulation graph policy (green) show that the dynamics model training and bisimulation metric is likely working here, while the additional errors induced by the corresponding cloned policy (red) show that here, the behavior cloning process induces a slight additional error. We also note that the value function errors are much smaller when the bisimulation metric is used (red / green) than when the Euclidean metric is used (blue / orange), providing empirical evidence for its theoretical benefits. Finally, on the right we see a very sensible looking value function plot, where values are higher as the policy winds its way out of the valley.


\section{Related Work}
\paragraph{Offline Reinforcement Learning}



In the past two years, there has been substantial interest in offline reinforcement learning and many algorithms have been developed to address the problem. In order to mitigate the issues of distribution shift and model exploitation, recent prior work in offline RL have explored incorporating many different types of conservatism in the optimization process. These approaches can be broadly grouped based on the type of conservatism they incorporate.

The first set of approaches use actor-critic methods \citep{wu2019behavior,kumar2019stabilizing,CRR,nair2020accelerating}, but incorporate policy constraints to limit the difference between the learned policy and the behavioral policy that collected the data in order to mitigate distribution shift during the Bellman updates and at test time. The second set of approaches use model-based RL \citep{yu2020mopo, morel}, but leverage uncertainty-aware dynamics models to perform Model-Based Policy Optimization (MBPO) \citep{mbpo} while deterring the policy from taking action with high model uncertainty. The third set of approaches add conservatism directly to the Q-function \citep{kumar2020conservative, yu2021combo} in order to optimize a lower-bound on the true value function and avoid extrapolating optimistically. Finally, an alternate approach attempts to filter out the best available transitions in the data for behavior cloning by learning an upper bound on the value function \cite{chen2020bail}.

\paragraph{Graphical Methods in Reinforcement Learning}
There have been recent prior works which leverage finite MDPs and their graphical representations in order to estimate value functions. The simplest take the highest-returning actions from a particular state \citep{blundell2016modelfree} or leverage differentiable memory to give a weighted combination of recent experiences \citep{neural_episodic_control}.
\citet{ETADP} gives a method of Exact then Approximate Dynamic Programming. The method quantizes the state space into variable-sized bins and treats the quantized MDP as finite. They solve this MDP via Value Iteration and then use these values to warm-start DDQN \citep{van2016deep}.
This method is close to ours, but assumes discrete action space, quantizes the state space, and does not leverage a dynamics model.

Another method, DeepAveragers \citep{shrestha2021deepaveragers}, constructs a finite MDP from the dataset and extends the optimal value function via kNN. Their theoretical analysis relies on assumptions on the Lipschitzness of Bellman backups that directly affects the value of a cost hyperparameter, while we use the properties of the bisimulation metric to guarantee our bounds. It also only works on problems with discrete actions and doesn't add to the dataset in any way.

Other methods, like \citet{Zhu*2020Episodic, HuEpsiodic}, use episodic memory techniques to organize the replay buffer, using averaging techniques and implicit planning to generalize from the replay buffer. However, they cannot plan novel actions to explicitly connect the dataset and are not designed with the offline setting in mind.


\section{Experiments}
\label{sec:experiments}

\begin{figure}[ht]
    \centering
    \includegraphics[width=\linewidth]{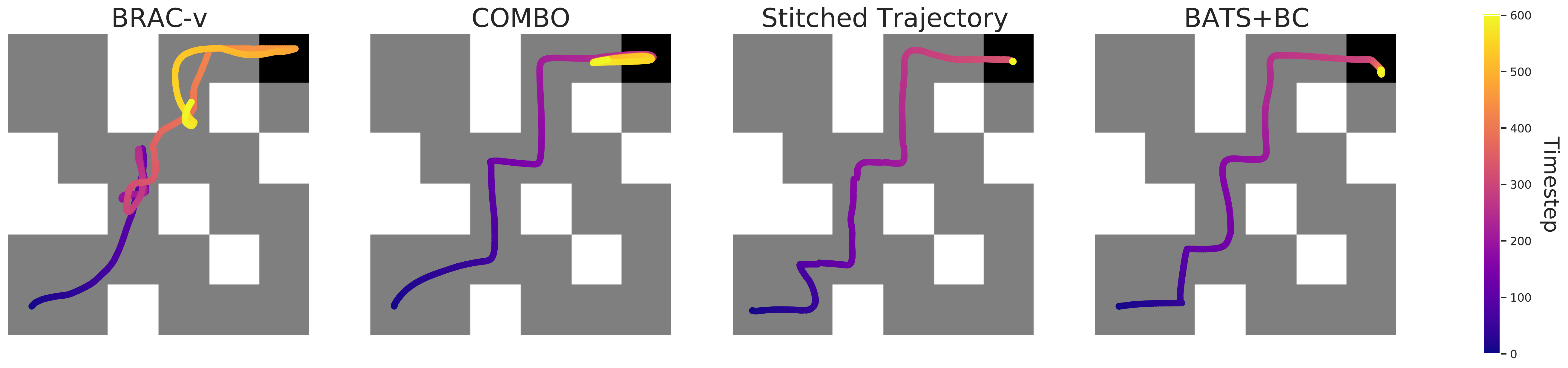}
    \caption{\textbf{Trajectories in maze2d-medium.} The two leftmost plots show the top baselines on this task. While BRAC-v is able to make it to the goal, it is apparent that the constraints imposed make it sub-optimal as it makes several loops throughout the trajectory. While COMBO is able to efficiently navigate to the goal, it too is unable to remain in the goal region.}
    \label{fig:medmaze_compare}
\end{figure}
\begin{table}[ht]
    \centering
    \begin{tabular}{| c | c | c | c | c | c | c | c | c |} \hline
        \textbf{Task} & \textbf{BATS+BC (Ours)} & \textbf{BC} &  \textbf{SAC} & \textbf{BRAC-v} & \textbf{BEAR} & \textbf{CQL} & \textbf{COMBO} \\ \hline
        umaze & \textbf{141.8} $\pm$ \textbf{4.4} & 3.8 & 88.2 & -16.0 & 3.4 & 5.7 & 76.2 $\pm$ 0.5 \\
        medium & \textbf{133.6} $\pm$ \textbf{11.6} & 30.3 & 26.1 & 33.8 & 29.0 & 5.0 & 74.8 $\pm$ 35.2 \\
        large & 107.7 $\pm$ 22.0 & 5.0 & -1.9 & 40.6 & 4.6 & 12.5 & \textbf{150.3} $\pm$ \textbf{22.0} \\ \hline
    \end{tabular}
    \caption{\textbf{D4RL Maze 2D.} The above shows undiscounted returns for each algorithm with the higest average bolded. For algorithms that we ran, we include the standard error. Results are averaged over three seeds, except for BATS+BC which was averaged over three seeds on each of the three graphs. \vspace{-6mm}}
    
    \label{tab:d4rl_results}
\end{table}

In this section we explore BATS experimentally. By planning in the approximate MDP, BATS can identify high-value sequences of actions. Unlike other state-of-the-art offline RL algorithms, BATS can also reason about which regions of the state space are important for the task at hand. We find that this is crucial for ``undirected'' datasets (i.e., datasets in which the data were collected without a specific reward function in mind).

In the following experiments we first use BATS to find a good policy in the approximate, finite MDP, and then learn a policy for the true, continuous MDP by behavior cloning on optimal trajectories from the approximate MDP. We also assume that we have access to the start state distribution, and we use this to go through the data and label additional states as start states, since many datasets have few starts. Once BATS is complete, data is collected by unrolling trajectories with the optimal policy. Because we find that not all logged trajectories were stitched to good regions of state space, we filter out any trajectory that does not meet a required value threshold. Then, a neural network policy that outputs the mean and variance of a normal distribution is learned on the collected data. 

To implement the algorithm, we rely on the \textit{graph-tool} library. We used compute provided by XSEDE~\citep{towns2014xsede} to run BATS three times for each task. In order to save on compute, we execute the BATS algorithm with no penalty and with relaxed hyperparameters. After, we perform simple grid search of hyperparameters by relabelling and deleting stitched transitions and re-running value iteration on the resulting MDP. For the penalty term, we use $L2$ distance scaled by a constant hyperparameter. For more details on this procedure see Appendix~\ref{app:hyperparameters}.

For baselines, we compare to methods reported in D4RL: off-policy SAC \cite{SAC}, CQL \cite{kumar2020conservative}, BRAC \cite{wu2019behavior}, BEAR \cite{kumar2019stabilizing}. Since D4RL does not report any model-based methods, we compare against the COMBO algorithm, which alters the CQL algorithm to include short model rollouts. We used our own implementation of COMBO, which we find gets comparable results to the results reported (see  Appendix~\ref{app:combo_implementation} for details). Final results can be seen in Table~\ref{tab:d4rl_results}.

\textbf{2D Maze Environments.} We evaluate our procedure on D4RL's \textit{maze2d} environments. While the baseline algorithms we compare against produce amazing results on the Mujoco control tasks, they struggle to match the performance of an expert policy on these relatively simple maze environments. We assert that a key reason for this is that the dataset is undirected. For each of the maze tasks, the dataset was collected by a PD controller that randomly wandered through the mazes. This presents a problem for algorithms which try to constrain the learned policy to be close to the behavioral policy, since there may be many instances where the behavioral policy performs an action that is in direct opposition to a specified goal. We see this concretely in the maze case, where most of the baseline policies are able to find their way to the goal, but start moving away from the goal once it is reached (see Figure~\ref{fig:medmaze_compare}). This happens because the policies are trained on a significant amount of data where the behavioral policy leaves the goal, but there are few to no data where the behavioral policy remains stationary in the goal. Even COMBO, which leverages a learned dynamics model, is unable to stay in the goal cell for the umaze and medium maze tasks.

BATS addresses this problem directly by filtering out any data that are unimportant for the task at hand. Training solely on the data seen by the optimal agent in the approximate MDP ensures that the policy for the continuous MDP is never trained on any disastrous actions or any data that are too far out of the policy's state distribution. At the same time, BATS can use the learned dynamics model to reason about how to stay stationary within the goal cell, as shown by Figure~\ref{fig:medmaze_compare}.

\vspace{-3mm}
\section{Conclusion}
\vspace{-3mm}
In this work, we presented an algorithm which eschews learning a deep value function by instead maintaining a finite MDP of transitions from the dataset and augmenting it with transitions planned using a learned model. The algorithm's performance is promising on low-dimensional tasks with theoretical support for its value function estimates. While stitching is hard on higher-dimensional tasks, we believe this work can be improved by incorporating low dimensional representation learning. Ideally this representation should be related to the bisimulation metric. Although we achieved promising results using the model architecture as described in \citet{zhang_bisim} on mountain car, we were unable to leverage the same model in our other experiments. We hope that new developments in learning bisimulation metrics will unlock additional potential in BATS. We also hope to formalize the equivalent algorithm for when transitions are stochastic, and we hope that this extension will help the algorithm generalize to more complex problems. 



\begin{acksection}
This material is based upon work supported by the National Science Foundation Graduate
Research Fellowship Program under Grant No. DGE1745016. Any opinions, findings, and
conclusions or recommendations expressed in this material are those of the author(s) and do
not necessarily reflect the views of the National Science Foundation.

This work used the Extreme Science and Engineering Discovery Environment (XSEDE), which is supported by National Science Foundation grant number ACI-1548562.
\end{acksection}
\newpage

{
\small
\bibliographystyle{abbrvnat}
\bibliography{references}
}

\newpage
\appendix

\section{The BATS Algorithm}
\label{s:subroutines}

Below we state the steps of the BATS algorithm. Algorithm~\ref{algo:bats} shows the main loop, while Algorithm~\ref{algo:subroutines} shows the subroutines.

\begin{algorithm}
\caption{The BATS Algorithm}
\label{algo:bats}
\begin{algorithmic}
\Procedure{BATS}{Offline Dataset $D = \bigcup_{j \in [M]}\{(s_{ji}, a_{ji}, s'_{ji}, r_{ji})\}_{i=1}^{t_j}$, Max Stitching Length $K$, Number of Iteration $n$, Number of Samples per Iteration $m$, Neighborhood Radius $\epsilon$, Planning Tolerance $\delta$, Discount Factor $\gamma$, Penalty Coefficent $c$, and Distance Metric $d$} \\
\State $M_0 = \langle \Statespace_0, \{\Actionspace^s_0\}_{s \in \Statespace_0}, \gamma, T_0, r_0, \rho_0\rangle$
\State \textrm{Learn dynamics estimate, $\tilde{T}$, and reward estimate, $\tilde{r}$ from $D$}
\For{$i = 0, 1, \ldots, (n - 1)$}
    \State $\hat{V}_i^*(\cdot), \pi_i(\cdot) \gets \textrm{valueIteration}(M_i)$
    \State neighbors $\gets \getNeighbors(\Statespace, \pi, M_i, \epsilon)$
    \State $M_{i + 1} \gets M_i$
    \For{$j=1, 2, \ldots, m$}
        \State $s \sim \mu_{M_i}(s \mid \pi_i)$
        \State $E \gets \getPotentialEdges(s, M_i, \text{neighbors}, j)$
        \ForAll{$(s, s') \in E$}
            \State actions = $\testEdge((s, s'), K, \delta)$
            \State $s'' = s$
            \ForAll{$a \in$ actions}
                \State $\Actionspace_{i + 1}^{s''} \geq \{a\} \cup \Actionspace_{i + 1}^{s''}$
                \If{$a$ is last action in actions}
                    \State $T_{i + 1}(s'', a) \gets \tilde{T}(s'', a)$
                    \State $r_{i + 1}(s'', a) \gets \tilde{r}(s'', a) - c d \left( \tilde{T}(s'', a), T(s'', a) \right)$
                    \State $s'' \gets \tilde{T}(s'', a)$
                    \State $\Statespace_{i + 1} \gets \{s''\} \cup \Statespace_{i + 1}$
                \Else
                    \State $T_{i + 1}(s'', a) \gets s'$
                    \State $r_{i + 1}(s'', a) \gets \tilde{r}(s'', a) - c d \left( \tilde{T}(s'', a), s' \right)$
                \EndIf
            \EndFor
        \EndFor
    \EndFor
\EndFor
\State $V(\cdot), \pi(\cdot) \gets \textrm{valueIteration}(M_n)$
\Return $M_n$
\EndProcedure

\end{algorithmic}

\end{algorithm}

\begin{algorithm}
\caption{Subroutines for BATS}
\label{algo:subroutines}
\begin{algorithmic}
\Function{$\getPotentialEdges$}{$s, t, \hat{M}, \text{neighbors}, j$}
\If{k == 0}
    \Return []
\EndIf
\State edges = []
\State vertexNeighbors $\gets$ $\text{neighbors}[t]$
\State MDPNeighbors $\gets \hat{M}[s, \hat{\Actionspace}_s]$
\ForAll{$n \in$ vertexNeighbors}
    \State edges += $\getFutureEdges$($s$, $n$, $\hat{M}$, $j$)
\EndFor
\ForAll{$n \in$ MDPNeighbors}
    \State successorNeighbors $\gets \text{neighbors}[t]$
    \ForAll{$n' \in$ successorNeighbors}
        \State neighbors += [($s$, $n'$)]
    \EndFor
    \State neighbors += $\getPotentialEdges(s, n, \hat{M}, \text{neighbors}, j - 1)$
\EndFor

\Return edges
\EndFunction
\Function{$\getFutureEdges$}{$s, n, \hat{M}, j$}
    \State MDPNeighbors $\gets \hat{M}[s, \hat{A}_s]$
    \State edges $\gets$ []
    \ForAll{$n' \in$ MDPNeighbors}
        \State edges += $[(s, n')]$
        \State edges += $\getFutureEdges(s, n', \hat{M}, j-1)$
    \EndFor

\Return edges
\EndFunction
\Function{$\testEdge$}{e, k, $\epsilon$}
    \State minDistance $\gets \infty$
    \State bestActions $\gets []$
    \For{i = 1:j}
        \State distance, actions $\gets \CEM(e, k)$
        \If{distance < minDistance}
            \State minDistance $\gets$ distance
            \State bestActions $\gets$ actions
        \EndIf
    \EndFor
    \If{minDistance < $\epsilon$}
        \Return []
    \EndIf
    
    \Return bestActions
\EndFunction
\end{algorithmic}
\end{algorithm}

\newpage
\section{Additional Explanation of Bisimulation}
\label{a:bisimulation}
In this work, we make use of \emph{bisimulation metrics} as introduced by \citet{givan_equivalence_2003} in order to guarantee accurate value function estimates in our stitched graph. Informally, bisimulation metrics compare states based solely on their expected future reward distribution and thus ignore information that does not affect the reward. Specifically, they are constructed in \citet{FernsFiniteMDP} as a generalization of the following equivalence relation:
\begin{definition}
$\sim$ is a bisimulation relation if $s\sim s'$ implies that $\forall a \in \Actionspace, r(s, a) = r(s', a)$ and $T(s, a) \sim T(s', a)$.
\end{definition}
This means that two states are \emph{bisimilar} (related via $\sim$) if they attain the same one-step rewards and their future state distributions also return the same under the same actions. Intuitively, this relation serves to ignore attributes of the state space which are not related to the returns attained executing actions.
 This relation and its derived equivalence classes group together the states in which the agent receives the same rewards under the same actions for an arbitrary choice of actions. However, the restriction that states are only similar if they give the same future rewards under any sequence of actions is quite strong--if an obviously bad action executed in a pair of otherwise bisimilar states gives different rewards which are both bad, those states will not be bisimilar. Any competent reinforcement learning algorithm will not take an obviously bad action, so the states will be indistinguishable for RL purposes. 

To rectify this, \citet{castro2019scalable} give a coarser bisimulation relation and associated metric which is based on the actions a particular policy would take:
\begin{definition}
The on-policy bisimulation relation $\onpolbisim$ for a stochastic policy $\pi$ is the strongest relation such that $s\onpolbisim s'$ iff $\sum_{a}\pi(a\mid s)r(s, a) = \sum_a \pi(a\mid s')r(s', a)$ and $\forall C \in S / \onpolbisim$, $P^\pi(C\mid s) = P^\pi(C \mid s')$ where $S / \onpolbisim$ is the partition induced by $\onpolbisim$ and $P^\pi(C\mid s) = \sum_{a \in \Actionspace}\left(\pi(a \mid s) \mathbf{1}[T(s, a) \in C]\right)$.
\end{definition}

These works also give metrics which relax these equivalence classes, preserving a notion of approximate bisimilarity. In our work, we use the \emph{on-policy bisimulation distance} from \citet{castro2019scalable}, which also gives a sampling-based algorithm for approximating such a metric, which we'll denote $d_\sim^\pi(\cdot, \cdot)$. As one might imagine, it turns out that this distance is closely related to the value function for $\pi$ by the following theorem from the paper:

\begin{theorem}[Theorem 3 from \citet{castro2019scalable}]
Given any two states $s, t \in \Statespace$ in an MDP $M$ and a policy $\pi$, $|V^\pi(s) - V^\pi(t)| \leq d_\sim^\pi(s, t)$.
\end{theorem}
This result gives us a metric over states for which the value function is $1$-Lipschitz continuous. This property allows us to unify the graphical perspective of stitching we take with the more traditional value function approach to RL.

There is an extensive literature on bisimulation, and further discussion, theory, and empirical investigation can be found in \cite{ferns_continuous, FernsFiniteMDP, zhang_bisim, castro2019scalable}.



\section{Proof of Theorem \ref{thm:value_function_bound}}
We prove theorem \ref{thm:value_function_bound} in this section. The proof proceeds by taking the infinite expansion of a value function and correcting for planning errors using Theorem \ref{thm:bisimulation}. We can do this for each transition which is added by BATS instead of being take from the dataset, allowing us to compare the upper and lower bounds with their true values.
\label{ss:value_function_proof}
\begin{proof}
WLOG, suppose that for a fixed $z \leq \ell$, $\pi^-(b_i) = a_i$ if $i \leq z$ and $\pi^-(b_i) \neq a_i$ if $z < i \leq \ell$. In other words, the $\pi^-$ chooses to take advantage of $z$ of the stitches made. Note that $z$ can possibly be $0$, but in this case the theorem holds trivially.

Let $T^-$ be the transition function in $M^-$, and let $s_0, s_1, \ldots$ be the infinite sequence of states that $\pi^-$ visits in $M^-$ starting from $s_0 = s$ and where $\hat{s}_{i} = T^-(\hat{s}_{i - 1}, \pi^-(\hat{s}_{i - 1}))$. Let $\tau$ be the ordered set hitting times of the states where $\pi^-$ uses a stitched transition plus $0$ and $\infty$, i.e. $\tau = \{0\} \cup \{t | t \in \mathbb{N}^+ \textrm{ s.t. } s_t \in \{b_i\}_{i = 1}^z\} \cup \{\infty\}$, and let $t_i$ be the $i^{th}$ sorted element of $\tau$.



We can expand the value function $V(s)$ as follows:
\begin{align*}
    \label{eq:v_expansion}
    V_M^{\pi^-}(s) &= \sum_{i=0, s_0 = s, s_i = T(s_{i - 1}, \pi^-(s_{i - 1}))}\gamma^i r(s_i, \pi^-(s_i))\\
    &= \sum_{t_i \in \tau}\left (\sum_{j = t_i, s_j = T(s_{j - 1}, \pi^-(s_{j - 1}))}^{j=t_{i + 1} - 1} \gamma^{j}r(s_j, \pi^-(s_j)) \right).
\end{align*}
Leveraging our mapping, $\phi$, and Theorem~\ref{thm:bisimulation}, note that
\begin{align*}
    V_M^{\pi^-}(s_{t_i}) &= r(s_{t_i}, \pi(s_{t_i})) + \gamma V_M^{\pi^-}(s_{t_i + 1}) \\
    &= r(s_{t_i}, \pi(s_{t_i})) + \gamma V(s_{t_i + 1}) + \gamma V_M^{\pi^-}(c_{t_i + 1}) - \gamma V_M^{\pi^-}(c_{t_i + 1}) \\
    &\leq r(s_{t_i}, \pi(s_{t_i})) - \gamma \norm{\phi(s_{t_i + 1}) - \phi(c_{t_i})} + \gamma V_M^{\pi^-}(c_{t_i}) \\
    &\leq r(s_{t_i}, \pi(s_{t_i})) - \gamma \epsilon_{t_i} + \gamma V_M^{\pi^-}(c_{t_i}) \\
\end{align*}
We can apply this inequality at each hitting time to get the below:
\begin{equation*}
    \label{eq:true_lower_bound}
    V_M^{\pi^-}(s) \geq \sum_{t_i \in \tau}\left(\gamma^{t_i}r(s_{t_i}, \pi^-(s_{t_i})) - \gamma^{t_i + 1}\epsilon_i +
    \sum_{j = {t_i +1}, s_{t_i} = c_{t_i}, s_j = T^-(s_{j - 1}, \pi^-(s_{j - 1}))}^{j=t_{i + 1} - 1}\gamma^{j}r(s_j, \pi^-(s_j))\right)
\end{equation*}
\begin{equation*}
    \label{eq:true_upper_bound}
    V_M^{\pi^-}(s) \leq \sum_{t_i \in \tau}\left(\gamma^{t_i}r(s_{t_i}, \pi^-(s_{t_i})) + \gamma^{t_i + 1}\epsilon_i +
    \sum_{j = {t_i +1}, s_{t_i} = c_{t_i}, s_j = T^-(s_{j - 1}, \pi^-(s_{j - 1}))}^{j=t_{i + 1} - 1}\gamma^{j}r(s_j, \pi^-(s_j))\right).
\end{equation*}
Note that, by construction, these lower and upper bounds equal the value functions in the MDPs $M^-$ and $M^+$, respectively.
\begin{equation}
    \label{eq:lower_value}
    V_{M^-}^{\pi^-}(s) = \sum_{t_i \in \tau}\left(\gamma^{t_i}r(b_{t_i}, \pi^-(b_{t_i})) - \gamma^{t_i + 1}\epsilon_i +
    \sum_{j = {t_i +1}, s_{t_i} = c_{t_i}, s_j = T^-(s_{j - 1}, \pi^-(s_{j - 1}))}^{j=t_{i + 1} - 1}\gamma^{j}r(s_j, \pi^-(s_j))\right)
\end{equation}
and
\begin{equation}
    \label{eq:upper_value}
    V_{M^+}^{\pi^-}(s) = \sum_{t_i \in \tau}\left(\gamma^{t_i}r(b_{t_i}, \pi^-(b_{t_i})) + \gamma^{t_i + 1}\epsilon_i +
    \sum_{j = {t_i +1}, s_{t_i} = c_{t_i}, s_j = T^-(s_{j - 1}, \pi^-(s_{j - 1}))}^{j=t_{i + 1} - 1}\gamma^{j}r(s_j, \pi^-(s_j))\right).
\end{equation}
Combining the above gives the desired result for our arbitrary $s$:
\begin{equation*}
    V_{M^-}^{\pi^-}(s) \leq V_M^{\pi^-}(s) \leq V_{M^-}^{\pi^-}(s).
\end{equation*}
\end{proof}
\section{Experiment Details}

\subsection{Hyperparameters and Training Procedure}
\label{app:hyperparameters}

\textbf{Dynamics Models.} To learn dynamics models, we use the architecture introduced by \citet{PETS}, and follow the procedure described in \citet{yu2020mopo}, making a few minor changes. Like \citet{yu2020mopo}, we train seven different dynamics models and take the best five based on a validation set of 1,000 points. Each model is a neural network with 4 hidden layers of size 200 and 2 heads at the end: one predicting mean and one predicting log-variance. These models are trained using batches of size 256 and using negative log likelihood as the loss. We use ReLU for our hidden activation function, and unlike \citet{yu2020mopo}, we do not use spectral normalization. Following their procedure, we use a validation set of one thousand points to select the best model to use after training.

\textbf{BATS.} For each of the experiments in Section~\ref{sec:experiments}, we perform BATS three times, using different learned dynamics models, to produce three different stitched MDPs. When using CEM to plan new stitches, we use the mean output of each member of the dynamics model and check if the $80^{th}$ quantile is under some planning threshold. This planning threshold was set to $0.425$ with for the mazes and $10$ (after normalization) for Mujoco tasks. All experiments imposed a restriction of $k=1$ number of actions that could be taken for stitching, except for halfcheetah which we set to $k=5$. Additionally, when forming the nearest neighbor graph for finding potential stitches, we consider neighbors up to $0.225$ away for umaze and medium maze and $0.15$ away for large maze. These were set as large as memory constraints would allow for. For Mujoco tasks we found it easier to instead use the 25 closest neighbors.

We assume for these experiments that we have access to the start state distribution. For Mujoco tasks we simply label the beginning of trajectories in the logged dataset as start states. Since there is only one trajectory for each of the maze tasks, we label every state that is in the support of the start state distribution as a start state. The large maze dataset does not contain possible start states for all cells. For cells in which there is not a start state in the dataset, we widen the distribution slightly so that enough starts are included.

For the maze tasks we attempt $50,000$ stitches every iteration, and we run BATS for $10$ iterations for umaze and medium maze and $20$ iterations for the large maze. For the Mujoco experiments, we attempt to make $5,000$ stitches ever iteration, and we run BATS for $40$ iterations.

To increase exploration for the stitches to consider, we apply Boltzmann exploration when selecting next actions to perform in the stitched MDP. That is, we select actions according to:
\begin{align*}
    \Prob(a | s) = \propto \exp \left( Q(s, a) / T \right)
\end{align*}
where $T$ is a temperature parameter, which we choose to set to $0.25$ for all experiments.

After running BATS, we searched for good hyperparameters by relabeling or removing stitched edges accordingly. The best found parameters are shown in Table~\ref{tab:hps}. When looking at the distribution of returns from trajectories in the resulting MDPs, there is a clear value for the returns that separates successfully stitched trajectories from those that were not able to be stitched to high value areas. As such, we only behavior clone on trajectories above $100$ (umaze), $200$ (medium maze), $300$ (large maze), $1,000$ (hopper and walker2d), and $4,000$ (halfcheetah). These thresholds were selected by inspection.

\begin{table}[]
    \centering
    \begin{tabular}{|c|c|c|c|c|c|c|} \hline
        & \multicolumn{3}{|c|}{\textbf{Mazes}} & \multicolumn{3}{|c|}{\textbf{Mujoco Mixed Tasks}} \\ \hline
         & \textbf{umaze} & \textbf{medium} & \textbf{large} & \textbf{hopper} & \textbf{walker2d} & \textbf{halfcheetah}\\ \hline
        Planning Error Threshold & \multicolumn{3}{|c|}{0.425} & 2.25 & 0.5 & 2 \\ \hline
        Penalty Coefficient & \multicolumn{2}{|c|}{20} & \multicolumn{2}{|c|}{10} & \multicolumn{2}{|c|}{50} \\ \hline
        Policy Layer Sizes & $64,64$ & $256,256$ & $256,256,256$ & \multicolumn{3}{|c|}{$256,256$} \\ \hline
        Batch Size & \multicolumn{6}{|c|}{256} \\ \hline
        Batch Updates & $10,000$ & \multicolumn{2}{|c|}{$20,000$} & \multicolumn{3}{|c|}{$10,000$} \\ \hline
    \end{tabular}
    \caption{\textbf{Table of BATS Hyperparameters.}}
    \label{tab:hps}
\end{table}

\textbf{Mountaincar.} 

For the mountain car example, we found that doing 20 iterations of stitching (each trying to make 100 stitches) was sufficient. We use the nearest 25 neighbors to determine which stitches can be made, and we allow for up to $k=5$ actions to be used when stitching between states. We found that smaller dynamics models were sufficient for this problem, and in particular, each member of the ensemble had 3 layers with 64 hidden units each. Lastly, we set the the temperature for Boltzmann exploration to $T=0.1$. We behavior clone using a policy network with two hidden layers with $256$ units each.

\subsection{Bisimulation Implementation}
\label{app:bisim_implementation}

The model architecture that we use for bisimulation is that of \citet{zhang_bisim}. That is, we have a network that takes state observations as inputs and outputs a latent representation, and we have a dynamics model that operates on this latent representation. For the encoder network, we use three hidden layers with 256, 128, and 64 units, respectively, and we set the latent dimension to be 6. Unlike the model in \citet{zhang_bisim}. however, we have one network that predicts both next transitions in bisimulation space and next rewards (the same dynamics model as described in Appendix~\ref{app:hyperparameters}). We also use the same loss function as described in \citet{zhang_bisim}. In particular, we draw batches of pairs of state observations and optimize according to
\begin{align*}
    J(\phi) = \left(\norm{z_i - z_j} - |r_i - r_j| - \gamma W_2 \left(\hat{P}(\cdot | \overline{z}_i, a_i),  \hat{P}(\cdot | \overline{z}_j, a_j)\right)\right)^2
\end{align*}
where $z_k, r_k, a_k$ are the latent encoding, the predicted reward, and the observed action for the $k^{th}$ sample, respectively. $\hat{P}$ is the learned dynamics model for the latent space, and we use a bar over $z$ to signify that we stop gradients. For more details, please refer to \citet{zhang_bisim}. Although in their work they iteratively update their model to reflect a changing policy, in our work we train with respect to a fixed policy. The on-policy nature of our training procedure resembles \citet{castro2019scalable}.

\subsection{COMBO Implementation}
\label{app:combo_implementation}

Because the original COMBO \citep{yu2021combo} paper did not include results on the maze environments and does not yet have a public implementation, we made our best attempt at reimplementing their method in order to properly compare results. For running COMBO, we based our dynamics model training on \citet{yu2020mopo} and used all of their associated hyperparameters. For the conservative Q-learning and policy learning components, we mostly followed the public implementation of \citet{kumar2020conservative}, but had to make some small tweaks to make it consistent with the descriptions in \citet{yu2021combo}.

For the COMBO hyperparameters, we did a grid search over conservative coefficient $\beta$ in $\{0.5, 1.0, 5.0\}$ and rollout length $h$ in $\{1, 5\}$ for all the maze tasks. We found the best parameters to be $h=1, \beta=1$ on umaze, $h=1, \beta=0.5$ on medium, and $h=1, \beta=0.5$ on large. For all the other hyperparameters, we followed the halcheetah parameters used in \citet{yu2021combo}. Specifically, 3-layer feedforward neural networks with 256 hidden units for Q-networks and policy, Q-learning rate $3.e-4$, policy learning rate $1.e-4$, $\rho(a|s)$ being the soft-maximum of the Q-values and estimated with log-sum-exp, $\mu(a | s) = \pi(a | s)$, and deterministic backups.

\end{document}